\documentclass[sigconf]{acmart}

\usepackage{CJKutf8}

\usepackage{array}
\usepackage{makecell}
\usepackage{float} 
\usepackage{bm}
\usepackage{graphicx}
\usepackage{booktabs}
\usepackage{multirow}
\usepackage{enumitem}
\setlist[itemize]{leftmargin=*}
\setlist[enumerate]{leftmargin=*}
\definecolor{lightgray}{RGB}{215,215,215}
\definecolor{lightblue}{RGB}{107,174,214}
\definecolor{bluu}{HTML}{ECF4FF}
\definecolor{blu}{RGB}{158,202,225}
\definecolor{myorange}{RGB}{2, 142, 2}
\usepackage{colortbl}  
\usepackage{color}
\usepackage{xcolor}
\usepackage[normalem]{ulem}
\useunder{\uline}{\ul}{}
\usepackage{subfigure}
\usepackage{wrapfig}
\usepackage{amsmath}

\newcommand{\sss}[1]{\subsubsection{\textbf{#1}}}
\newcommand{\ie}{\emph{i.e., }}
\newcommand{\eg}{\emph{e.g., }}

\definecolor{projectblue}{HTML}{0077B5}

\newcommand{\projectlinks}{%
  \mbox{%
    \href{https://OpenFinArena.com}
      {\textcolor{projectblue}{\underline{Project Website}}}%
    \hspace{1.2em}%
    \textcolor{black!45}{|}%
    \hspace{1.2em}%
    \href{https://github.com/transcend-0/FinDeepIndicator}
      {\textcolor{projectblue}{\underline{GitHub}}}%
    \hspace{1.2em}%
    \textcolor{black!45}{|}%
    \hspace{1.2em}%
    \href{https://huggingface.co/datasets/OpenFinArena/FinDeepIndicator}
      {\textcolor{projectblue}{\underline{Hugging Face}}}%
  }%
}

\makeatletter
\patchcmd{\@mktitle@iii}
  {\par\bigskip}
  {%
    \par\vspace{0.5em}%
    {%
      \centering
      \normalfont
      \sffamily
      \small
      \mdseries
      \projectlinks
      \par
    }%
    \vspace{0em}%
  }
  {}
  {%
    \PackageError{project-links}
      {Could not patch the ACM title block}
      {The installed acmart version may have changed.}%
  }
\makeatother

\setcopyright{acmlicensed}
\copyrightyear{2027}
\acmYear{2027}
\acmDOI{XXXXXXX.XXXXXXX}
\acmConference[XXX]{XXXXXX}{XX, XX}{XX, XX}
\acmISBN{978-1-4503-XXXX-X/2018/06}

\begin{document}


\title{FinDeepIndicator: Benchmarking Deep Research Agents in End-to-End Financial Indicator Construction}


\author{
Chaoqun Yang$^{1}$, Fengbin Zhu$^{1*}$, Xinyu Lin$^{1*}$, Long Bai$^{1,4}$,\\Xiaoluan Liu$^{2}$, Ke-Wei Huang$^{3}$, Roger Zimmermann$^{1}$, Tat-Seng Chua$^{1}$
}
\thanks{$^{*}$Corresponding authors}
\affiliation{
\institution{
$^1$School of Computing, National University of Singapore; 
$^2$China Economics and Management Academy, Central University of Finance and Economics;
$^3$Asian Institute of Digital Finance, National University of Singapore;
$^4$Institute of Computing Technology, Chinese Academy of Sciences
}
\country{}
}
\email{chaoqun@yang.email.cn, fengbin@nus.edg.sg, xylin1028@gmail.com, bailong@ict.ac.cn, xiaoluanliu@email.cufe.edu.cn, dishkw@nus.edu.sg, dcsrz@nus.edu.sg, dcscts@nus.edu.sg}

\renewcommand{\shortauthors}{Chaoqun Yang et al.}

\begin{abstract}
Financial indicators are essential tools for transforming raw financial data into interpretable measures for various downstream tasks, such as valuation, risk assessment, and economic analysis. However, existing financial benchmarks largely focus on answer-level accuracy and often assume that relevant data are already provided, leaving the assessment of the intermediate process of indicator construction underexplored. In this work, we propose \textbf{FinDeepIndicator}, the first benchmark dedicated to evaluating Deep Research (DR) agents in end-to-end financial indicator construction. Specifically, FinDeepIndicator evaluates DR agents across four stages in indicator construction: formula specification, data collection, indicator calculation, and answer generation, and covers fundamental, technical, and macroeconomic indicators organized into 21 fine-grained sub-categories.
It contains 3,350 curated question-answer (QA) pairs derived from both U.S. and Chinese markets, 10 years of historical financial data, and 800 listed companies.
Extensive experiments on search-equipped Large Language Models (LLMs) and DR agents show that, while LLMs generally perform well in formula specification, their accuracy drops substantially during data retrieval and numerical execution. DR agents consistently outperform search-equipped LLMs, yet remain unreliable in realistic financial analysis settings.
These findings provide insights for developing more capable and trustworthy DR agents in finance.

\end{abstract}





\maketitle

\section{Introduction}

Financial indicators are structured quantitative summaries that bridge raw financial data and financial decision-making~\cite{murphy1999technical, mishkin2007economics, penman2010financial}. They transform heterogeneous financial signals (\eg financial statements, asset prices, and macroeconomic time series) into compact numerical measures for valuation, profitability analysis, and macroeconomic interpretation. Representative examples include price-to-earnings ratios, moving averages, volatility measures, inflation rates, and interest-rate spreads. Because these indicators are widely used and frequently updated, they form a fundamental interface between financial knowledge and computational reasoning. A financial agent should therefore not only interpret indicator semantics, but also construct indicators on demand: given a natural-language query, it should infer the precise underlying formula, obtain the requisite raw data from the web, execute the computation, and yield a verifiable result.

However, existing benchmarks do not adequately evaluate this capability. First, there is no benchmark specifically designed for financial indicator construction; even when financial indicators appear in existing benchmarks, their coverage is limited and often incidental within broader financial question answering tasks~\cite{shah2022flue,zhu2024benchmarking,guo2025fineval,xie2024finben}. Second, most benchmarks assume that relevant documents, tables, or numerical inputs are already provided in context, thereby neglecting the data collection stage entirely, which fails to reflect real-world conditions~\cite{chen2021finqa,chen2022convfinqa,zhu2021tat,islam2023financebench,chen2024fintextqa,reddy2024docfinqa,lai2025sec,choe2025hierarchical,choi2025finagentbench}. Third, even in benchmarks involving search or tool use, evaluation is typically restricted to final-answer correctness, without assessing the intermediate process of indicator construction, which undermines the reliability of the generated results~\cite{bigeard2025finance,hu2026finsearchcomp,shen2025finsearch}. 

\begin{figure}[t]
    \centering
    \setlength{\abovecaptionskip}{0cm}
    \setlength{\belowcaptionskip}{-0.5cm}
    \includegraphics[width=1\linewidth]{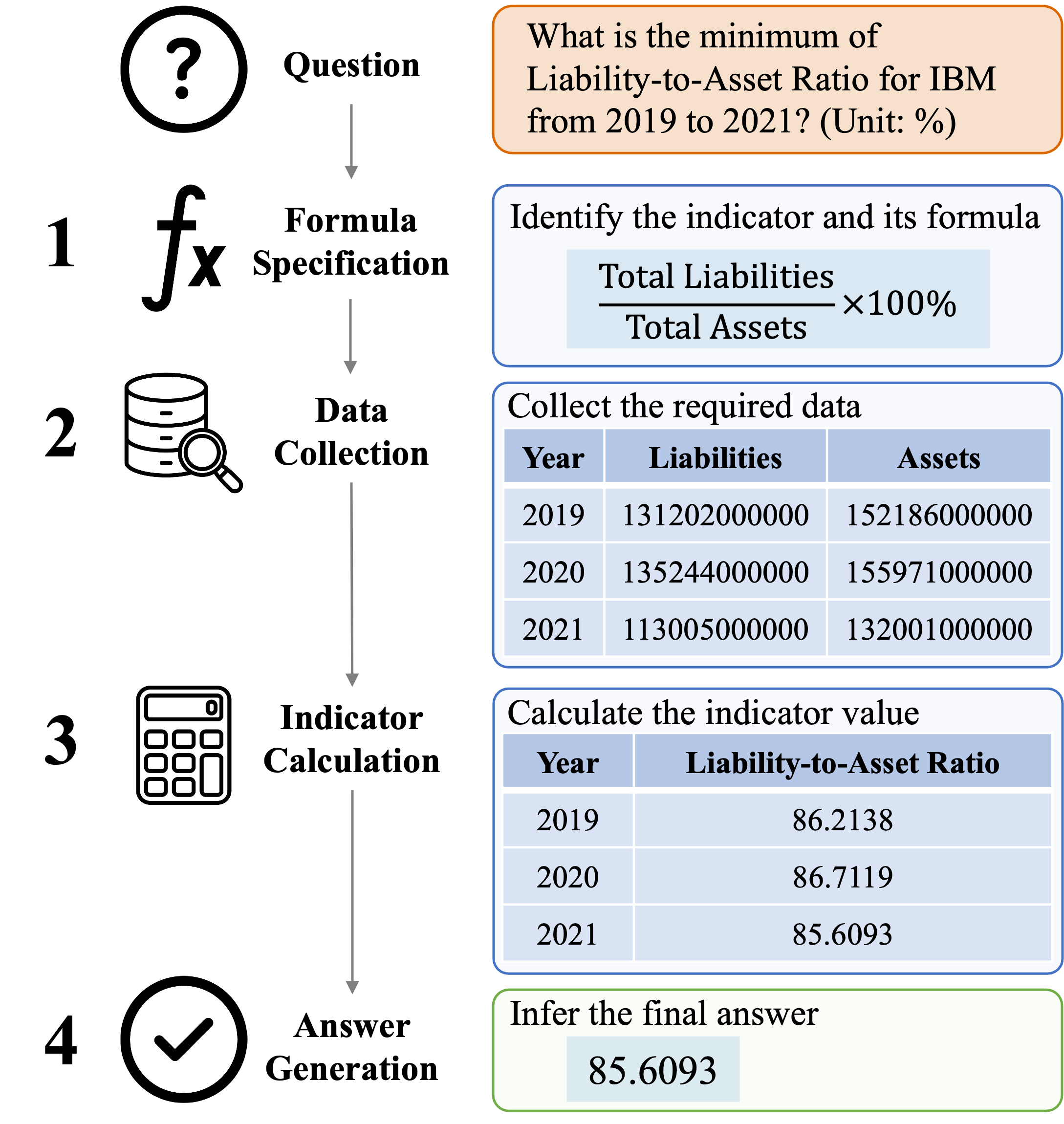}
    \caption{An example of financial indicator construction. 
    }
    \label{fig:case}
\end{figure}

To bridge this gap, we aim to evaluate Deep Research (DR) agents in end-to-end financial indicator construction. Figure~\ref{fig:case} illustrates a sample of end-to-end financial indicator construction.
The query asks for the minimum Liability-to-Asset Ratio of IBM from 2019 to 2021. Solving this problem requires multiple interconnected steps: the agent must first correctly identify the target indicator and its underlying formula, \ie \textit{Total Liabilities / Total Assets $\times$ 100\%}; it must then retrieve the required financial statement values for each year, align the corresponding liability and asset items, compute the yearly indicator values, and finally perform the specified temporal aggregation to obtain the minimum value. Errors may arise at any stage of this process, including incorrect indicator interpretation, inaccurate data retrieval, numerical computation mistakes, or inappropriate aggregation. 
Consequently, evaluating only the final answer provides an insufficient basis for assessing a model's financial reasoning capability and fails to reveal its specific limitations.

To address these challenges, we propose a process-level evaluation framework for end-to-end financial indicator construction. The framework decomposes the indicator construction process into four stages: \emph{Formula Specification}, \emph{Data Collection}, \emph{Indicator Calculation} and \emph{Answer Generation}. \emph{Formula Specification} evaluates whether an agent correctly identifies the financial concept and mathematical definition. \emph{Data Collection} measures whether the agent can retrieve relevant and accurate financial information. \emph{Indicator Calculation} evaluates numerical correctness under tolerance-aware matching to account for rounding and reporting variations. \emph{Answer Generation} assesses whether the agent correctly synthesizes intermediate results to infer the final answer. This process-level evaluation enables fine-grained analysis of agent capabilities and failure modes beyond conventional answer-only evaluation.

\begin{figure}[t]
    \centering
    \setlength{\abovecaptionskip}{0cm}
    \setlength{\belowcaptionskip}{-0.5cm}
    \includegraphics[width=0.95\linewidth]{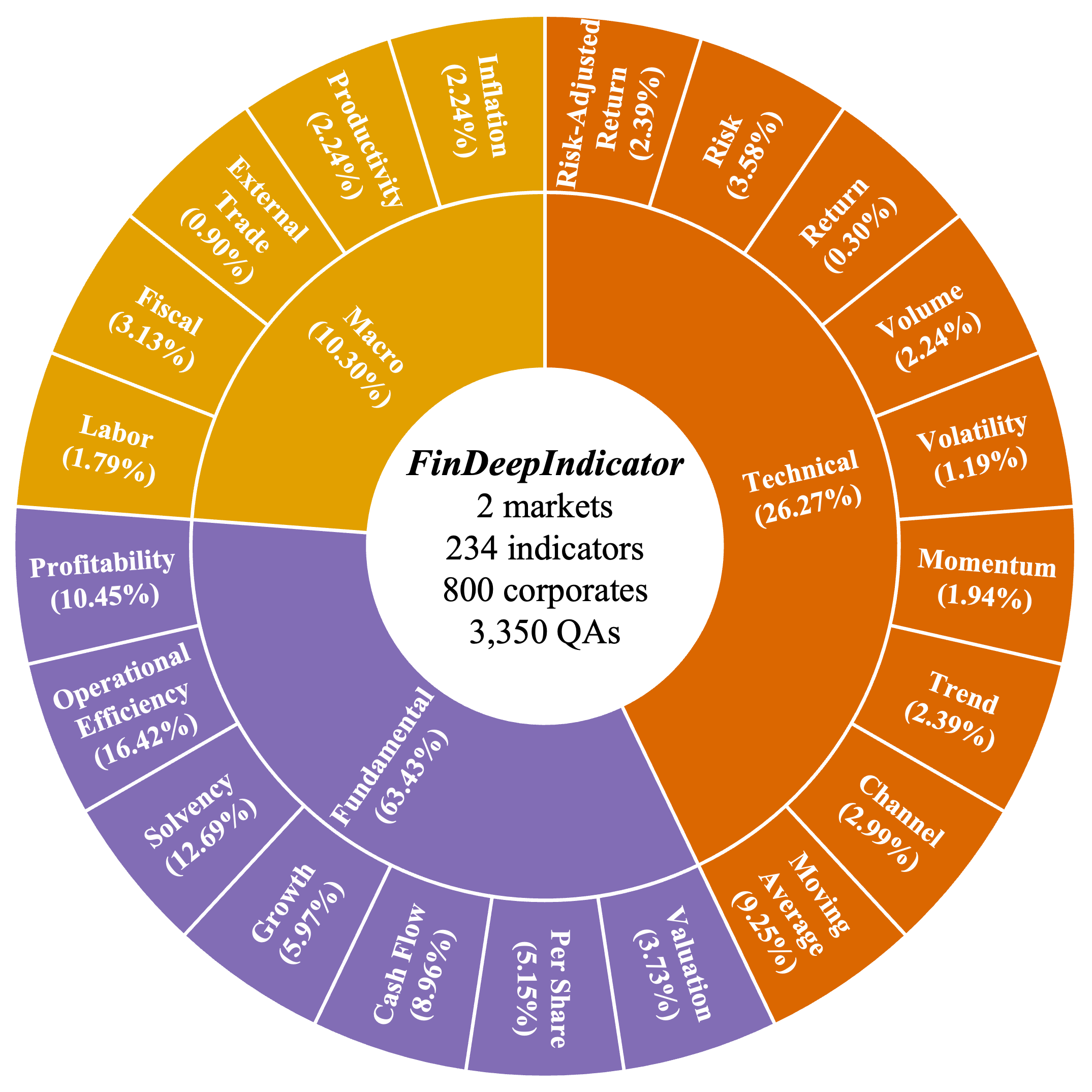}
    \caption{An overview of FinDeepIndicator.}
    \label{fig:overview}
\end{figure}

To support systematic benchmark construction, we collect and organize 234 commonly used financial indicators into three major categories and 21 fine-grained sub-categories, as shown in Figure~\ref{fig:overview}. The categories include fundamental, technical, and macroeconomic indicators, covering financial concepts such as profitability, valuation, momentum, volatility, inflation, and labor. For each indicator, we curate its standardized name, mathematical formula, required raw data, and executable calculation program, providing a structured foundation for automatic question generation, answer verification, and process-level evaluation. Based on the proposed evaluation framework and collected indicators, we construct \textbf{FinDeepIndicator}, the first benchmark specifically designed for assessing DR agents in end-to-end financial indicator construction. The benchmark spans the U.S. and Chinese markets and incorporates 10 years of historical data from 800 publicly listed companies. Overall, FinDeepIndicator contains 3,350 carefully curated question-answer pairs that require agents to complete the full construction workflow, including indicator understanding, raw data collection, intermediate indicator calculation, and final answer generation.

With FinDeepIndicator, we conduct extensive experiments to evaluate search-equipped LLMs and DR agents in end-to-end financial indicator construction. The key findings include: 1) Even the best-performing system achieves only approximately 40\% final-answer accuracy, despite most models exceeding 70\% in formula accuracy. This reveals a substantial gap between understanding financial indicators and executing their end-to-end construction, likely because models struggle with underlying raw data retrieval and precise data processing. 2) Two of the three strongest DR agents achieve higher final-answer accuracy in the U.S. market than in the Chinese market, by over 9.6\% and 3.1\%, respectively, while the remaining agent performs similarly across markets, indicating a modest but model-dependent cross-market gap that may arise from differences in data accessibility, source structure, and reporting conventions. 3) Macroeconomic indicators are the most challenging category, with final-answer accuracy below 30\% for most models, particularly for external-trade, fiscal, labor, and productivity indicators, likely because they require integrating heterogeneous sources and aligning reporting frequencies, geographic definitions, and temporal conventions.

In summary, this paper makes the following key contributions: 
\begin{itemize}
    \item To assess models' fine-grained capabilities in end-to-end financial indicator construction, we propose a \textbf{process-level evaluation framework} that decomposes the construction process into four distinct stages: \emph{Formula Specification}, \emph{Data Collection}, \emph{Indicator Calculation}, and  \emph{Answer Generation}. 
    \item To systematically capture the diversity and complexity of the financial indicator construction task, we design a comprehensive \textbf{financial indicator taxonomy} comprising 3 major categories, 21 sub-categories, and 234 distinct financial indicators.
    \item Building upon the proposed evaluation framework and indicator taxonomy, we construct \textbf{FinDeepIndicator}, the first benchmark specifically designed to assess end-to-end financial indicator construction, comprising 3,350 QA pairs derived from two major financial markets and 10 years of historical financial data from 800 listed companies.
    \item We systematically evaluate state-of-the-art search-equipped LLMs and DR agents, revealing limited end-to-end performance, model-dependent gaps across markets, particular difficulty with macro indicators, and data collection as the primary bottleneck.
 \end{itemize}

\section{FinDeepIndicator Benchmark}

\subsection{Benchmark Construction}

\begin{figure}[t]
    \centering
    \setlength{\abovecaptionskip}{0.1cm}
    \setlength{\belowcaptionskip}{-0.5cm}
    \includegraphics[width=1\linewidth]{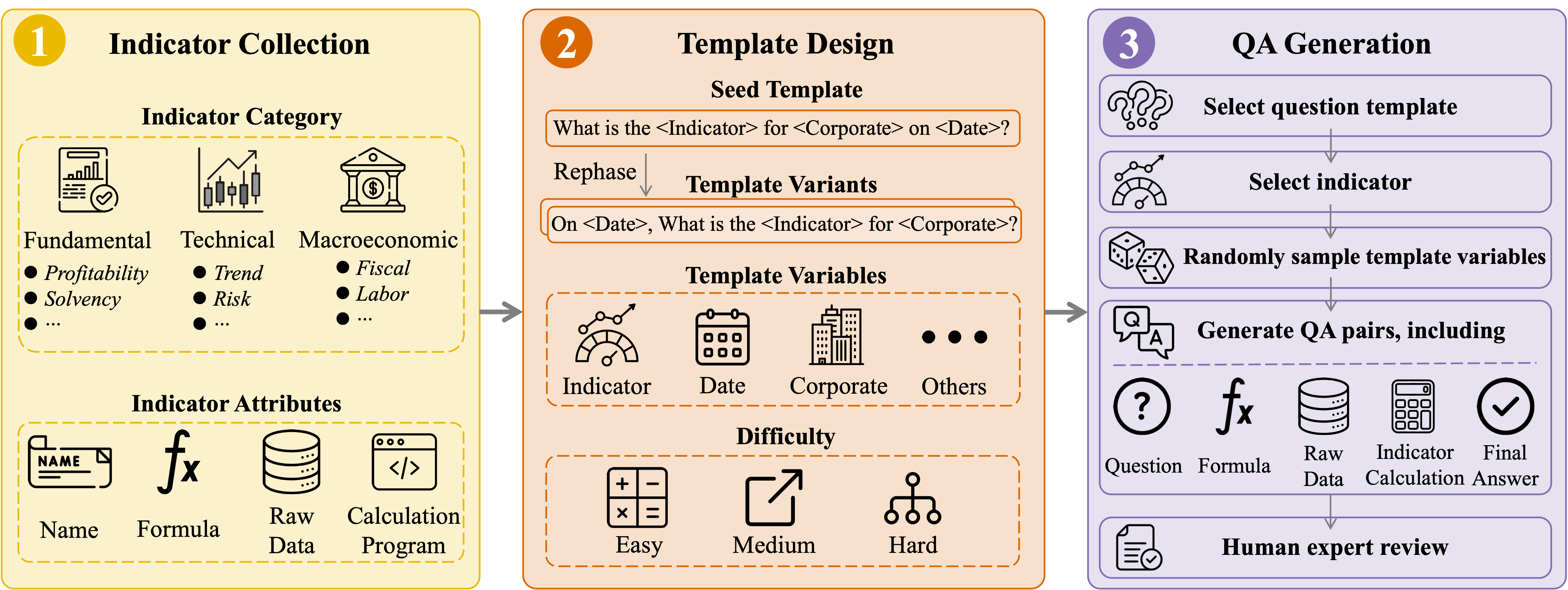}
    \caption{Construction pipeline of FinDeepIndicator.
    }
    \label{fig:pipeline}
\end{figure}

Figure~\ref{fig:pipeline} illustrates the construction pipeline of FinDeepIndicator. 
The pipeline comprises three stages: 1) \emph{Indicator Collection}, gathers widely used derived financial indicators and organizes their names, formulas, required raw data, and calculation procedures. 
2) \emph{Template Design}, develops indicator-specific templates with configurable variables and three difficulty levels. 
3) \emph{QA Generation}, instantiates these templates through random variable sampling to produce QA pairs accompanied by detailed intermediate processes.

\sss{Indicator Collection.}

We first collect financial indicators widely used in practical financial analysis, then organize them into 3 major categories and 21 sub-categories, forming a comprehensive financial indicator taxonomy, as shown in Figure~\ref{fig:overview}. 
In total, we obtain 234 distinct indicators, and the complete list of indicators is provided in Appendix~\ref{app:indicator}.
For each indicator, we document its name, mathematical formula, raw data, and calculation program. This structured metadata serves two purposes. 
First, it ensures that all generated questions have well-defined ground-truth computation procedures. Second, it supports process-level evaluation by making it possible to separately verify the formula, data, intermediate calculation, and final answer.

\sss{Template Design.}

We design a set of broadly applicable and linguistically diverse templates to cover common analytical requests involving financial indicators. These templates support tasks such as computing an indicator for a company, comparing indicator values across years, and identifying the maximum or minimum value within a specified period.

To construct the templates, we first group indicators by type and design shared templates for indicators within the same type. Each template is further associated with five surface variants that express the same underlying analytical intent in different natural-language forms, thereby increasing linguistic diversity while maintaining generation controllability. In addition, each template contains several placeholder variables, \eg indicator, company, date, conditional logic, etc. During QA generation, these placeholders are instantiated with values sampled from predefined variable pools.

When designing templates, we divide them into three difficulty levels: \emph{Easy}, \emph{Medium} and \emph{Hard}. \emph{Easy} questions require direct calculation of the target indicator from the collected raw data. \emph{Medium} questions require one-hop analysis in addition to indicator computation, such as selecting the maximum, minimum, or year-over-year change after computing a sequence of values. \emph{Hard} questions require multi-hop analysis, where the model must combine indicator computation with more complex comparison, aggregation, filtering, or temporal reasoning.
In total, we retain 170 templates.

\sss{QA Generation.}

Given the indicator metadata and question templates, we instantiate benchmark examples by exhaustively covering template variants and applicable indicators, while randomly sampling the remaining placeholder variables. Specifically, for each template variant, we enumerate all indicators to which the template is applicable. We then randomly sample remaining placeholder variables from predefined variable pools, such as company, date, and other task-specific variables. This design ensures systematic coverage over both linguistic variants and financial indicators, while maintaining diversity in entities, temporal settings, and task-specific conditions.

Each generated instance contains a natural-language question, the reference formula, the required raw data, the intermediate indicator calculation, and the final numerical answer.
Because each template variant can be paired with many applicable indicators and each indicator can be instantiated with many combinations of entities, dates, and reasoning operators, the resulting candidate pool contains billions of potential QA instances. From this pool, we generate 3,350 representative QA examples by enumerating all template variants and applicable indicators, and sampling valid combinations of the remaining variables.

\sss{Quality Control.}
We maintain the high quality of FinDeepIndicator by implementing a rigorous quality-control process throughout its construction and generation stages, including,

\begin{itemize}
    \item \textbf{Comprehensive indicator taxonomy.}
    We develop a comprehensive taxonomy covering diverse financial indicators. For each indicator, we record its definition, required inputs, calculation procedure, and applicability conditions. For example, profit margin requires revenue to be greater than zero, return on assets requires positive total assets, and return on equity requires available and non-zero shareholder equity.
    \item \textbf{Rigorous question design.}
    We additionally impose question-level constraints during template construction and instantiation. These constraints regulate date ranges, entities, comparison groups, and other query parameters to ensure that each question is unambiguous, answerable, and supported by available data.
    \item \textbf{Expert quality inspection.}
    After QA generation, we conduct random-sampling inspections with human experts to further validate dataset quality. The sampled QA instances are reviewed for question clarity, data availability, formula correctness, and calculation correctness. Instances containing invalid parameter settings or inconsistent computations are regenerated.
\end{itemize}

\subsection{Evaluation Protocol}

\sss{Process-Level Answer Extraction}

We evaluate model responses at multiple stages of the indicator construction process, including formula specification, raw data collection, indicator calculation, and final answer generation. This process-based evaluation provides a fine-grained analysis of model capabilities.

Model responses are typically free-form and exhibit substantial variation in format. We first use an LLM (\eg DeepSeek-V4-Flash) to parse each response into four structured components: the indicator formula, the raw data table, the indicator calculation table, and the final answer. The raw data and indicator calculation components are further normalized into CSV-style tables to enable consistent and fine-grained evaluation. 

To verify the reliability of the automatic parsing procedure, we conduct a criterion-based expert review following established evaluation practices~\cite{van2019best,elangovan2024considers}. Specifically, we randomly sample 100 model responses and invite three domain experts with financial expertise to independently assess whether each parsed component is correct and complete (See Appendix~\ref{app:expert_review_criteria} for expert review criteria). As reported in Table~\ref{tab:human_check}, the inter-annotator agreement for the four extracted components is consistently high, with all Gwet's AC1 scores exceeding 0.6. Moreover, the expert acceptance rates are all above 0.9, indicating that the LLM-based parsing procedure reliably converts free-form model responses into structured representations suitable for subsequent evaluation.

\begin{table}[t]
    \setlength{\abovecaptionskip}{0cm}
    \setlength{\belowcaptionskip}{0cm}
    \centering
    \caption{Inter-annotator agreement and expert validation results for response parsing and formula evaluation.
    For ``Formula Evaluation'', the entry in the ``Acceptance Rate'' row reports (winning rate, advantage probability).}
    \setlength{\tabcolsep}{1.5mm}
    \small
    \begin{tabular}{r|c|c|c|c|c}
    \toprule
    ~ & \makecell[c]{\textbf{Formula} \\ \textbf{Parse}} & \makecell[c]{\textbf{Data} \\ \textbf{Parse}} & \makecell[c]{\textbf{Indicator} \\ \textbf{Parse}} & \makecell[c]{\textbf{Answer} \\ \textbf{Parse}} & \makecell[c]{\textbf{Formula} \\ \textbf{Evaluation}} \\ \hline
    \makecell[c]{\textbf{Gwet's} \\ \textbf{AC1}} & 0.86 & 0.92 & 0.83 & 0.91 & 0.71 \\ \hline
    \makecell[c]{\textbf{Acceptance} \\ \textbf{Rate}} & 0.95 & 0.96 & 0.98 & 0.94 & (1.0, 0.86) \\
    \bottomrule
    \end{tabular}
    \label{tab:human_check}
\end{table}

\sss{Evaluation Metrics}

We evaluate four stages of the indicator construction process: formula specification, data collection, indicator calculation, and final answer generation.

\begin{itemize}
\item \textbf{Formula Specification Evaluation}.
We evaluate formula correctness based on semantic and mathematical equivalence rather than exact string matching, since the same financial indicator may be expressed using different but equivalent formulations. Specifically, we adopt an LLM-as-a-Judge protocol that assigns each parsed formula a three-level score according to whether it preserves the intended financial meaning, required variables, mathematical operations, and relevant domain conventions. The detailed scoring criteria are provided in Appendix~\ref{app:expert_annotation_rubric}. To assess the reliability of the LLM-based evaluator, we further conduct the Alt-test~\cite{calderon-etal-2025-alternative}. Three human experts independently evaluate the consistency between each parsed formula and the correct formula. As reported in Table~\ref{tab:human_check}, the Gwet's AC1 scores exceed 0.6, indicating substantial agreement among the human annotators. Moreover, the winning rate surpasses 0.5, passing the statistical test. Additionally, with an advantage probability exceeding 0.8, our results provide strong evidence that LLM annotation serves as a viable substitute for human evaluator.
\item \textbf{Data Collection Evaluation}.
We evaluate the collected raw data by comparing the generated data table with the corresponding reference data table. Both tables are represented in CSV format, where rows denote entities and columns denote the required fields, such as year and company. After aligning the rows and columns, we perform a cell-level comparison. For numerical entries, a cell is considered correct when the difference between the generated value and the reference value falls within a predefined tolerance (\eg 0.01, 0.005, or 0.001). This tolerance-aware comparison accommodates minor discrepancies caused by numerical precision, rounding, or unit conversion while still identifying missing or incorrect values. The data collection accuracy is then computed as the number of correctly matched cells divided by the total number of evaluated cells in the reference table.
\item \textbf{Indicator Calculation Evaluation}.
We evaluate the correctness of intermediate indicator calculations using the same table-based procedure. Specifically, the indicator calculation results are normalized into a CSV-style table and aligned with the corresponding reference table. Each numerical cell is evaluated using the predefined tolerance threshold, and the indicator calculation accuracy is calculated as the proportion of correctly matched cells among all evaluated cells. This metric measures whether the model correctly applies the indicator formula to the collected data and produces the expected intermediate indicator results.
\item \textbf{Answer Generation Evaluation}.
We evaluate the final answer using tolerance-aware numerical matching against the reference answer. A generated answer is considered correct if its numerical difference from the reference value is within the predefined tolerance, in which case the answer accuracy is assigned a value of 1. Otherwise, it is considered incorrect and assigned a value of 0. This evaluation accommodates minor discrepancies caused by rounding, numerical precision, unit conversion, or output formatting while distinguishing substantively incorrect answers.
\end{itemize}

\section{Experiments}

In this section, we conduct extensive experiments to systematically evaluate the capability of current LLMs and DR agents on financial indicator construction. Specifically, we aim to answer the following research questions:

\begin{itemize}
    \item \textbf{RQ1}: How capable are current LLMs and agents in performing end-to-end financial indicator construction?
    \item \textbf{RQ2}: How does the performance vary across different financial markets?
    \item \textbf{RQ3}: How does task difficulty affect model performance in financial indicator construction?
    \item \textbf{RQ4}: Which categories of financial indicators are more challenging for current models?
    \item \textbf{RQ5}: How does model performance change when the formula and raw data are provided?
    \item \textbf{RQ6}: What are the common error patterns made by current models during financial indicator construction?
\end{itemize}

\subsection{Evaluated Methods}
We evaluate two representative settings on FinDeepIndicator:

\begin{itemize}
    \item \textbf{LLM with Search (Search).} The LLM performs a single web search and then generates the final answer based on the retrieved information.
    \item \textbf{Deep Research Agent (Agent).} 
    Since existing DR systems differ substantially in implementation, making unified comparison difficult, we adopt a minimal DR agent based on the ReAct paradigm~\cite{yao2022react}. The agent iteratively reasons, invokes tools, and observes the results, with access to search and Python tools.
\end{itemize}

Evaluated LLMs include Qwen3.6-Max, Claude-Sonnet-4.6, Gemini-3-Flash, Deeepseek-V4-Flash, Grok-4.3, GPT-5, GPT-5-Mini. The equipped search API is Google Search.

\subsection{Overall Results (RQ1)}

\begin{figure}[t]
    \centering
    \setlength{\abovecaptionskip}{0cm}
    \setlength{\belowcaptionskip}{-0.5cm}
    \includegraphics[width=1\linewidth]{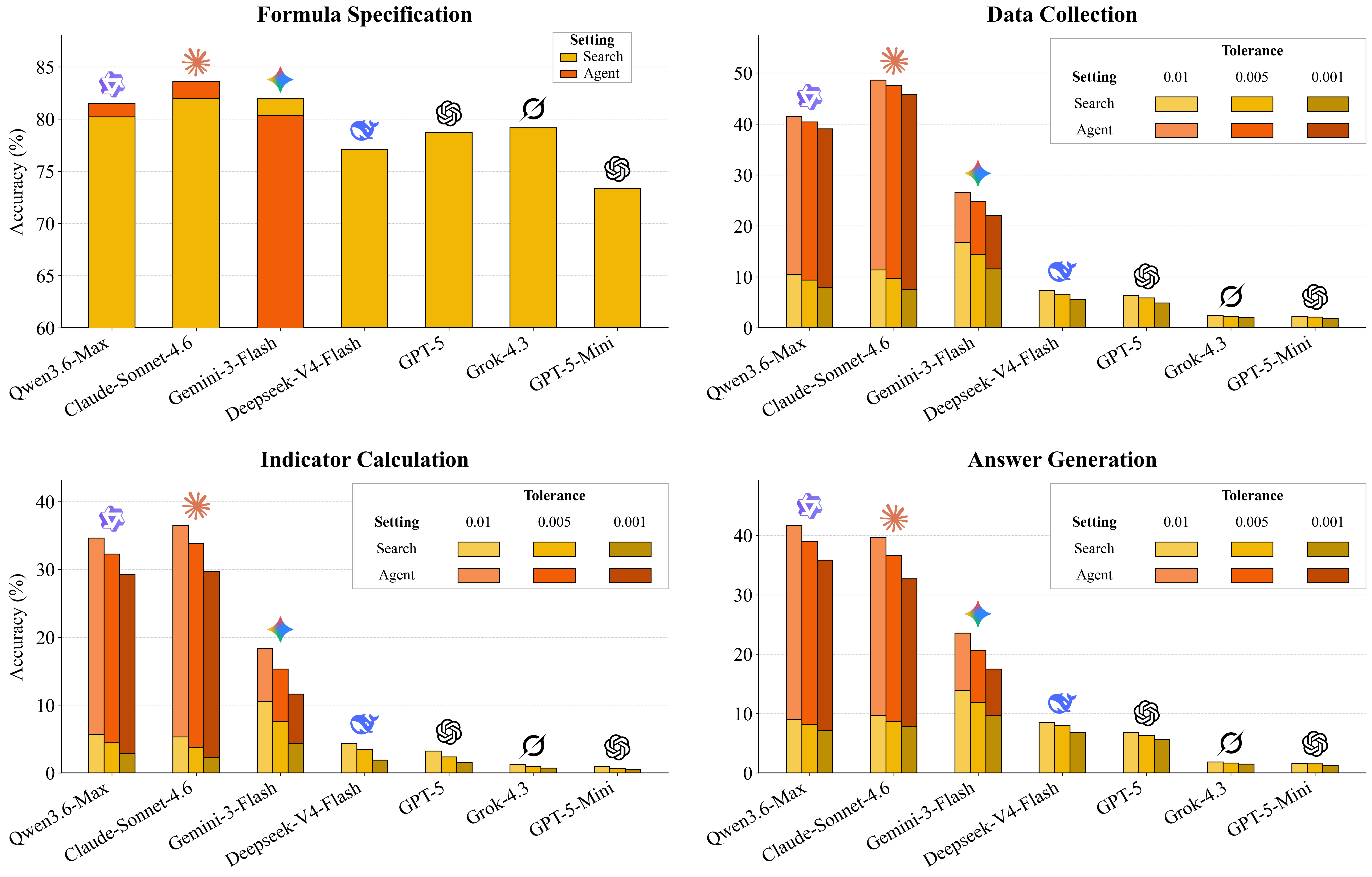}
    \caption{Overall performance of different models on FinDeepIndicator. Due to budget constraints, we evaluate the agent setting only on the three models that achieve the best performance under the search setting.
    }
    \label{fig:accuracy_overall}
\end{figure}

Figure~\ref{fig:accuracy_overall} presents the overall performance of different LLMs and agents across the four stages of financial indicator construction, from which we can observe the following:

\begin{itemize}
    \item \textbf{All methods show limited end-to-end construction ability.} Even the best-performing method achieves only around 40\% accuracy on final answers. Performance decreases further under stricter numerical tolerances. These results indicate that reliable end-to-end financial indicator construction remains challenging for current methods.
    \item \textbf{Agentic gains are substantial but model-dependent.} Agent-based settings generally outperform their search-based counterparts, particularly in data collection, indicator calculation, and final answer generation. However, the magnitude of improvement varies considerably across models. Gemini-3-Flash achieves the strongest performance in the search-only setting, but its agent-based variant provides a much smaller improvement than those of Qwen3.6-Max and Claude-Sonnet-4.6. This suggests that financial indicator construction depends not only on agents' retrieval capabilities but also on other agentic capabilities, such as planning and tool use.
    \item \textbf{A substantial gap exists between financial knowledge understanding and task execution.} Most methods achieve relatively high accuracy in formula specification, generally exceeding 70\%, which suggests that understanding financial concepts and identifying the corresponding formulas are not the main bottlenecks. In contrast, performance drops substantially in data collection and indicator calculation, revealing a clear gap between understanding financial knowledge and translating it into executable actions.
    \item \textbf{Data collection is the primary bottleneck.} The most significant performance drop is observed after the data collection stage, where accuracy decreases by approximately 40\% compared to formula specification. This indicates that data collection may be the most critical capability to improve in current models.
\end{itemize}

\subsection{Cross-Market Analysis (RQ2)}

\begin{figure}[t]
    \centering
    \setlength{\abovecaptionskip}{0cm}
    \setlength{\belowcaptionskip}{-0.5cm}
    \includegraphics[width=1\linewidth]{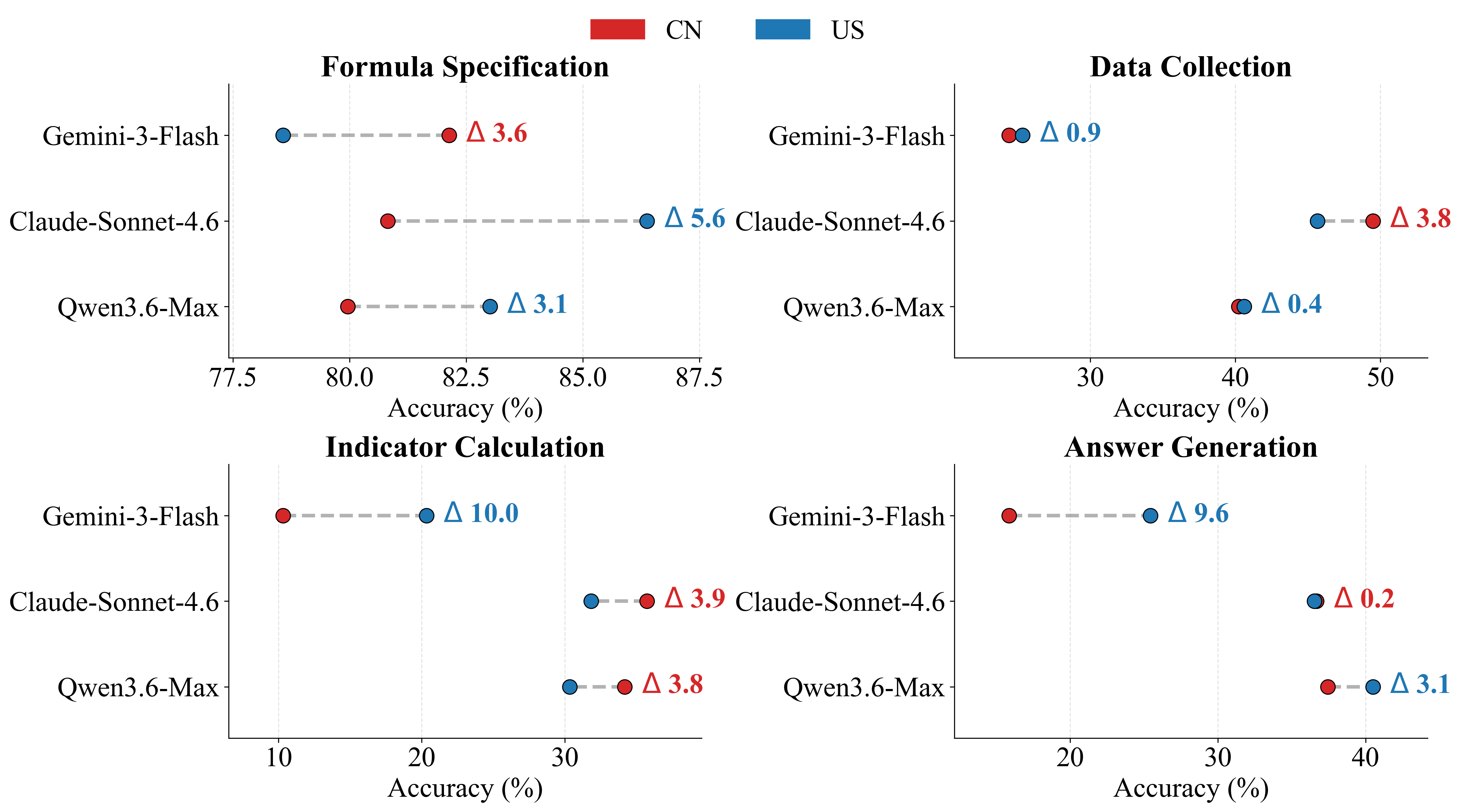}
    \caption{Performance comparison on agent setting between U.S. and Chinese financial markets.}
    \label{fig:accuracy_by_market}
\end{figure}

Figure~\ref{fig:accuracy_by_market} compares the performance of the three strongest LLM agents on FinDeepIndicator tasks drawn from the Chinese and U.S. markets. We make the following observations:

\begin{itemize}
    \item \textbf{Chinese-market tasks are slightly more challenging in terms of end-to-end performance.} In final answer generation, both Gemini-3-Flash and Qwen3.6-Max achieve higher accuracy on the U.S. market, while Claude-Sonnet-4.6 performs almost identically across the two markets. Overall, these results suggest that Chinese-market tasks are slightly more difficult at the end-to-end level. However, no consistent market-level advantage can be observed across all intermediate stages, as the relative performance varies by model and stage.
    \item \textbf{Different agents exhibit distinct market preferences.} Gemini-3-Flash performs substantially better on the U.S. market in indicator calculation and final answer generation, whereas Claude-Sonnet-4.6 achieves better performance on the Chinese market in data collection and indicator calculation. 
    This suggests that cross-market performance depends on each agent's retrieval and reasoning capabilities. Such stage-wise analysis can further identify agent-specific strengths and weaknesses, enabling targeted improvements or specialized agent collaboration across different construction stages.
\end{itemize}

\subsection{Impact of Task Difficulty (RQ3)}

\begin{figure}[t]
    \centering
    \setlength{\abovecaptionskip}{0cm}
    \setlength{\belowcaptionskip}{-0.5cm}
    \includegraphics[width=1\linewidth]{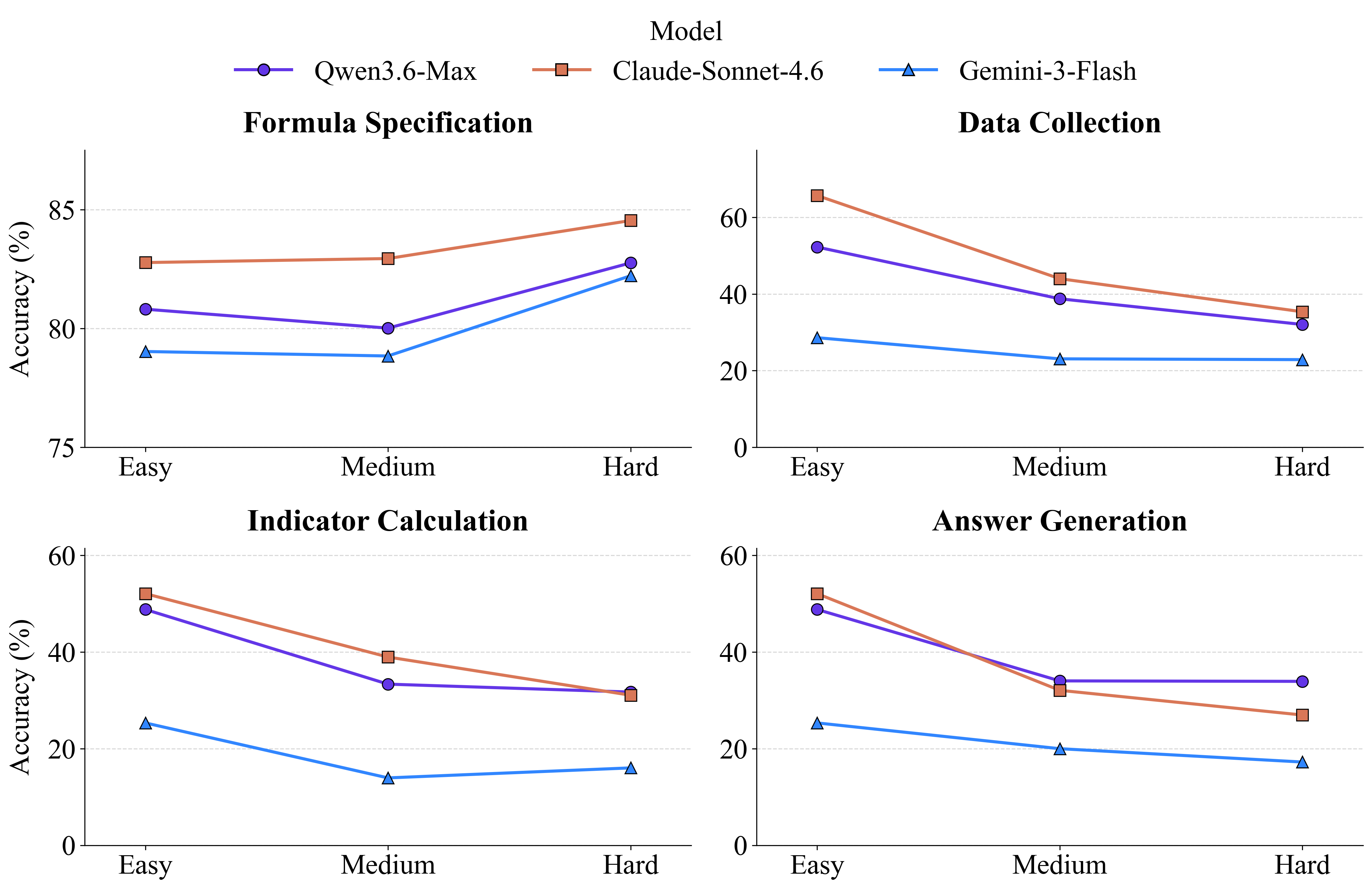}
    \caption{Performance comparison on agent setting under different task difficulty levels.}
    \label{fig:accuracy_by_difficulty}
\end{figure}

Figure~\ref{fig:accuracy_by_difficulty} compares the performance of the three strongest LLM agents across easy, medium, and hard instances in FinDeepIndicator. We make the following observations:

\begin{itemize}
    \item \textbf{End-to-end performance decreases as task difficulty increases.}
    All three agents achieve their highest final-answer accuracy on easy instances, followed by a considerable decline on medium and hard instances. Claude-Sonnet-4.6, for example, drops from approximately 52\% on easy tasks to below 30\% on hard tasks. This result demonstrates that current agents remain sensitive to increases in construction difficulty.
    \item \textbf{Task difficulty primarily affects data collection and indicator calculation.} 
    As task difficulty increases, performance declines in both stages, while formula specification remains comparatively stable. This suggests that greater logical complexity mainly increases the amount and difficulty of required data, the computational burden of indicator construction, and the reasoning needed for answer generation. The robustness of formula specification across difficulty levels further supports our earlier finding that financial knowledge understanding is not the primary challenge for current methods.
    \item \textbf{Different agents exhibit different levels of robustness to increasing task difficulty.} Claude-Sonnet-4.6 performs best on easy instances but experiences a continuous decline as the task difficulty increases. In contrast, Qwen3.6-Max maintains nearly unchanged final-answer accuracy from medium to hard instances and becomes the strongest model on the hard subset. Gemini-3-Flash performs consistently below the other two agents, although its indicator-calculation accuracy slightly recovers from medium to hard instances. This indicates that model rankings can change as tasks become more difficult.
\end{itemize}







\subsection{Performance Across Financial Indicator Sub-Categories (RQ4)}

\begin{figure*}[t]
    \centering
    \setlength{\abovecaptionskip}{0cm}
    \setlength{\belowcaptionskip}{0cm}
    \includegraphics[width=1\linewidth]{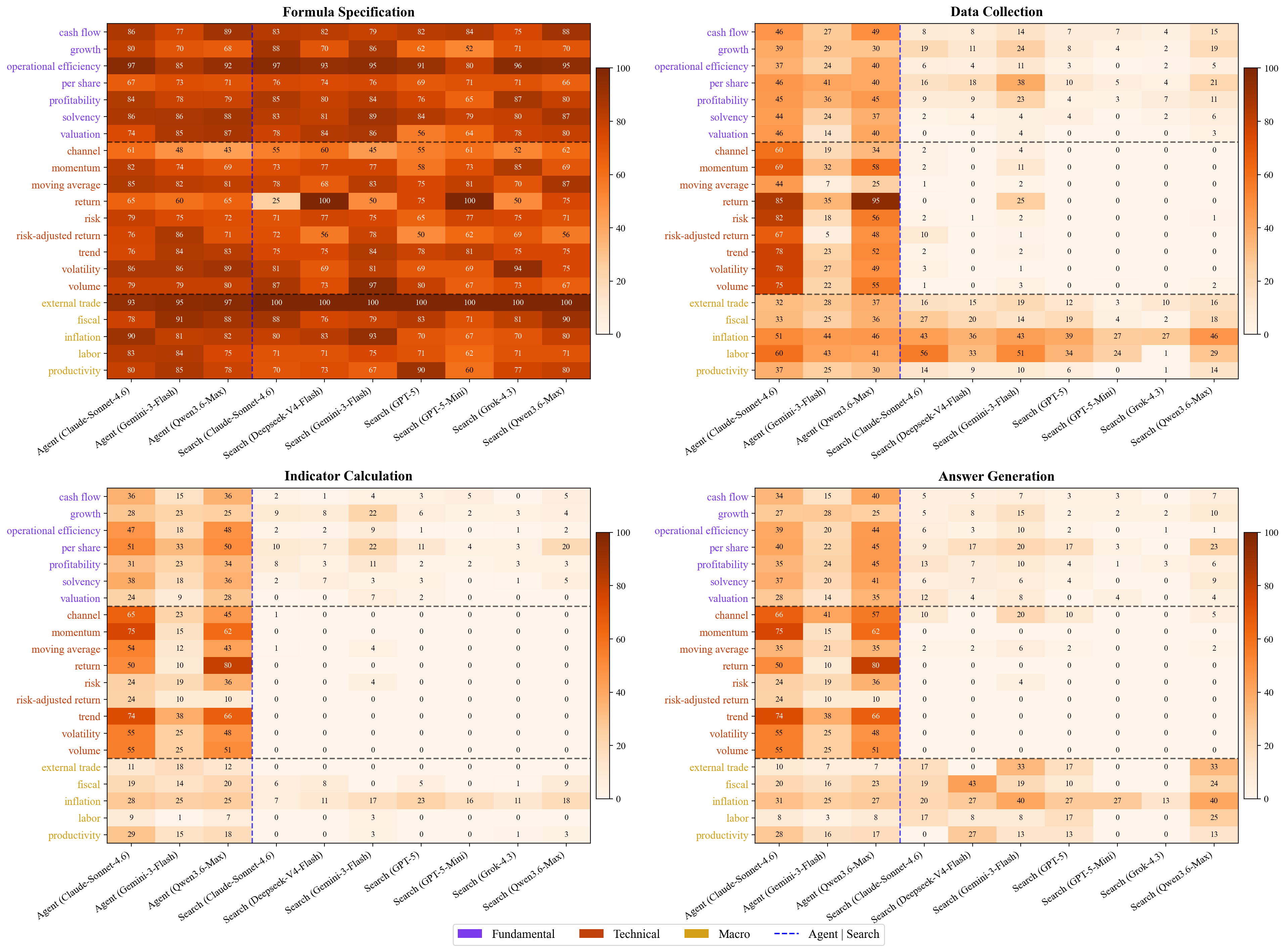}
    \caption{Fine-grained performance analysis across different financial indicator sub-categories.}
    \label{fig:accuracy_by_sub_category}
\end{figure*}

Figure~\ref{fig:accuracy_by_sub_category} presents a fine-grained comparison across indicator sub-categories, from which we make the following observations:

\begin{itemize}
    \item \textbf{Macro indicators show the weakest downstream performance.} Although formula accuracy remains high, indicator-calculation and answer-generation accuracy are generally low for external-trade, fiscal, labor, and productivity indicators. This is because macro tasks often involve different reporting frequencies, geographic scopes, seasonal adjustments, and temporal aggregation rules.
    \item \textbf{Agent performance varies substantially across technical indicator sub-categories.} Agents perform relatively well on channel, momentum, return, and trend indicators, but substantially worse on risk and risk-adjusted-return indicators. This gap likely arises because the latter require precise historical windows and statistical estimation.
    \item \textbf{Sub-category differences primarily emerge in downstream execution rather than formula understanding.}
    Formula accuracy remains consistently high across indicator sub-categories and models, indicating that current LLMs generally understand the definitions of diverse financial indicators. In contrast, data collection, indicator calculation, and answer generation accuracy vary substantially across sub-categories. This further confirms that the main difficulty lies in obtaining and processing the required data rather than identifying the appropriate formula.
    \item \textbf{No single agent dominates every indicator sub-category.} Claude-Sonnet-4.6, Qwen3.6-Max, and Gemini-3-Flash exhibit different strengths across indicator types. Qwen3.6-Max performs particularly well on several technical indicators, whereas Claude-Sonnet-4.6 is competitive across many fundamental and macroeconomic tasks. This variation indicates that indicator-specific retrieval and reasoning capabilities play an important role in overall model performance.
\end{itemize}

\subsection{Ablation Study on Data Collection (RQ5)}

\begin{figure}[t]
    \centering
    \setlength{\abovecaptionskip}{0cm}
    \setlength{\belowcaptionskip}{-0.5cm}
    \includegraphics[width=1\linewidth]{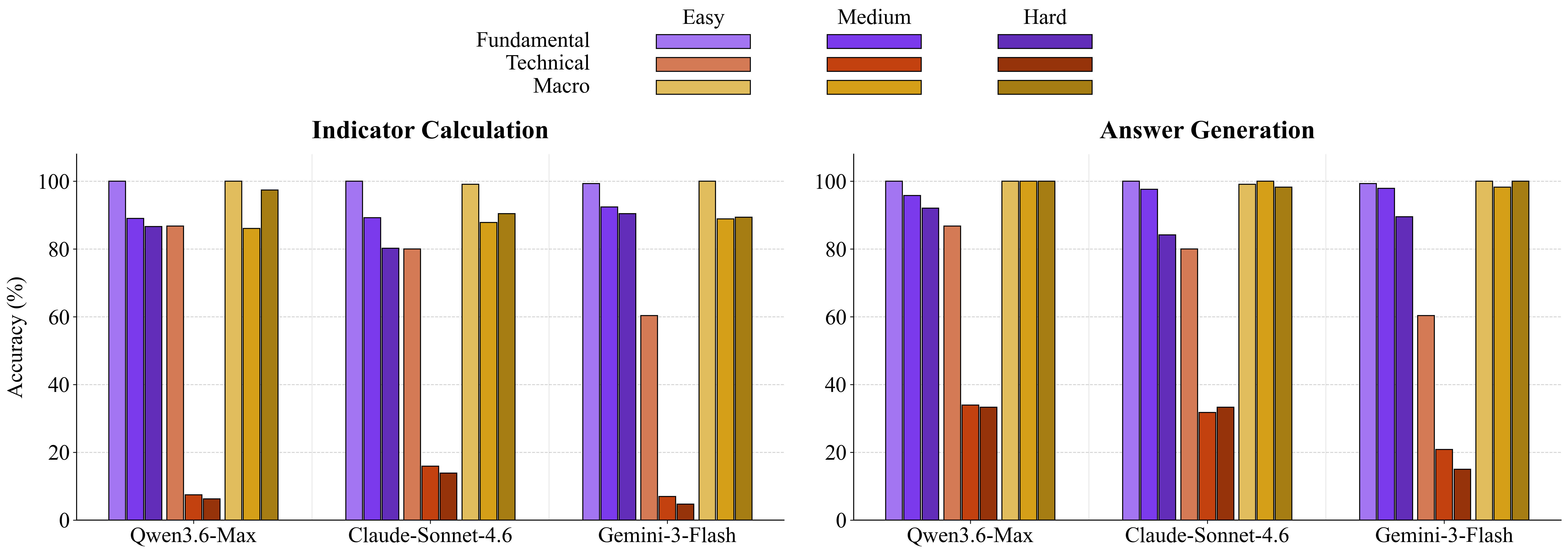}
    \caption{Performance comparison on agent setting when raw data is directly provided in context.}
    \label{fig:accuracy_given_data}
\end{figure}

Figure~\ref{fig:accuracy_given_data} reports model performance when the required formula and raw data are directly provided, thereby removing the need for formula specification and data collection. 
We make the following observations:

\begin{itemize}
    \item \textbf{The ablation study further confirms that data collection is the primary bottleneck.}
    Once the required raw data is provided, all three agents achieve substantially higher indicator-calculation and answer-generation accuracy than in the end-to-end setting. For fundamental and macroeconomic indicators, performance generally exceeds 80\% and frequently approaches 100\% across models and difficulty levels. This result reinforces our earlier finding that the main challenge lies in locating, extracting, aligning, and normalizing the required financial data, rather than in formula understanding or numerical computation.
    \item \textbf{Technical indicators remain challenging even when the data is provided.}
    In contrast to fundamental and macroeconomic indicators, technical indicators show substantially lower accuracy, especially on medium and hard instances. While easy technical instances can be solved with moderate to high accuracy, performance drops sharply as difficulty increases. 
    This indicates that technical indicators require more than correct data retrieval: they often involve complex temporal operations, rolling-window computation, sequential transformations, and precise handling of historical observations.
    \item \textbf{Task difficulty remains associated with performance after removing data collection.} 
    For both fundamental and technical indicators, performance continues to decline as task difficulty increases, with a substantially larger drop observed for technical indicators. In contrast, performance on macro indicators remains broadly stable across difficulty levels. Technical indicators typically involve more complex calculations and therefore remain challenging even when the required data are provided. In contrast, macro indicators generally involve simpler calculations, suggesting that their primary challenge lies in data collection rather than indicator calculation.
\end{itemize}

\subsection{Error Analysis (RQ6)}

\begin{table}[t]
    \centering
    \setlength{\abovecaptionskip}{0cm}
    \setlength{\belowcaptionskip}{0cm}
    \caption{Distribution of errors.}
    \label{tab:error_analysis}
    \setlength{\tabcolsep}{1mm}
    \begin{tabular}{llrr}
        \toprule
        \textbf{Error Aspect} & \textbf{Error Type} & \textbf{Percentage} & \textbf{Total} \\
        \midrule
        \multirow{2}{*}{\textit{Formula}}
            & Completely incorrect & 3.8\% & \multirow{2}{*}{8.6\%} \\
            & Partially incorrect  & 4.8\% & \\
        \midrule
        \multirow{4}{*}{\textit{Data}}
            & Incorrect data retrieval     & 50.1\% & \multirow{4}{*}{87.7\%} \\
            & Missing critical fields       & 19.8\% & \\
            & Incorrect time window         & 10.2\% & \\
            & Incorrect unit conversion     & 7.6\%  & \\
        \midrule
        \textit{Calculation}
            & Numerical calculation error   & 2.7\% & 2.7\% \\
        \midrule
        \textit{Logic}
            & Incorrect answer target       & 1.0\% & 1.0\% \\
        \bottomrule
    \end{tabular}
\end{table}

To better understand the limitations of current models, we randomly sample 100 failed cases from each of the three agent settings, resulting in 300 cases in total, and annotate the first error that occurs in each case. As summarized in Table~\ref{tab:error_analysis}, these errors fall into four categories: \textit{Formula}, \textit{Data}, \textit{Calculation}, and \textit{Logic}.
\begin{itemize} 
    \item \textbf{Formula errors.} Formula errors account for 8.6\% of all failures, including completely incorrect formulas (3.8\%) and partially incorrect formulas (4.8\%). The latter typically involves missing or substituted accounting items, or using an ending balance instead of the required average of beginning and ending balances. 
    \item \textbf{Data errors.} Data errors dominate the failures, accounting for 87.7\%. They include incorrect data extraction (50.1\%), missing critical fields (19.8\%), incorrect time windows (10.2\%), and unit-conversion errors (7.6\%). Common issues include selecting the wrong table entry and mixing reporting periods. 
    \item \textbf{Calculation errors.} Calculation errors account for 2.6\% of failures. They mainly result from arithmetic mistakes, premature rounding, or accumulated precision errors.
    \item \textbf{Logic errors.} Logic errors account for 1.0\% of failures. The model returns the wrong answer target, such as reporting an intermediate indicator value instead of the requested count. 
\end{itemize} 
\section{Related Work}

\subsection{Deep Research Agents}

Deep Research (DR) agents extend retrieval-augmented language models from single-pass retrieval to long-horizon workflows that involve task decomposition, iterative search, tool use, and evidence-grounded synthesis~\cite{huang2025deep, zhang2025deep}. This paradigm builds on agent frameworks such as ReAct~\cite{yao2022react}. Search-R1~\cite{jin2025search} and R1-Searcher~\cite{song2025r1} further train models to determine when and how to retrieve information during multi-step reasoning, although they mainly target question answering.
Recent systems, including OpenAI Deep Research~\cite{openai2025introducing}, Gemini Deep Research~\cite{gemini2025deep}, and Qwen Deep Research~\cite{qwen2025deep}, integrate planning, iterative retrieval, evidence management, and citation-grounded report generation. Accordingly, evaluation has expanded from retrieval and generation quality to the reliability of the overall research process.
General agent benchmarks evaluate complementary capabilities. GAIA~\cite{mialon2024gaia} covers reasoning, browsing, multimodal understanding, and tool use; FRAMES~\cite{krishna2025fact} focuses on factuality and multi-source reasoning; and BrowseComp~\cite{wei2025browsecomp} evaluates persistent web search. DR-specific benchmarks assess long-form outputs, including DeepResearch Bench~\cite{du2025deepresearch}, ReportBench~\cite{li2025reportbench}, and DEER~\cite{han2025deer}. However, most emphasize general-domain research or final-report quality rather than domain-specific workflows involving formula identification, raw-data collection, numerical transformation, and aggregation.

Financial indicator construction requires an agent to identify the correct formula, retrieve data for the relevant entities and periods, and perform the calculation. 
FinDeepIndicator therefore separately evaluates \emph{Formula Specification}, \emph{Data Collection}, \emph{Indicator Calculation}, and \emph{Answer Generation}, enabling process-level diagnosis.

\subsection{Financial Benchmarks}

Financial benchmarks broadly cover domain understanding, evidence-grounded numerical reasoning, and agentic evaluation in open information environments.
FLUE~\cite{shah2022flue} and CFLUE~\cite{zhu2024benchmarking} evaluate financial language tasks, while FinEval~\cite{guo2025fineval} and FinBen~\cite{xie2024finben} assess broader financial knowledge, reasoning, forecasting, and decision-making.
FinQA~\cite{chen2021finqa}, ConvFinQA~\cite{chen2022convfinqa}, and TAT-QA~\cite{zhu2021tat} focus on arithmetic reasoning over financial tables and text. Later benchmarks extend to long documents, open-book question answering, and retrieval-based settings, including FinanceBench~\cite{islam2023financebench}, FinTextQA~\cite{chen2024fintextqa}, DocFinQA~\cite{reddy2024docfinqa}, SEC-QA~\cite{lai2025sec}, LOFin~\cite{choe2025hierarchical}, and FinAgentBench~\cite{choi2025finagentbench}. These benchmarks generally provide the relevant evidence or restrict retrieval to a predefined corpus.
Recent benchmarks evaluate financial agents in open environments. Finance Agent Benchmark~\cite{bigeard2025finance} studies tool-augmented research; FinSearchComp~\cite{hu2026finsearchcomp} and FinSearchBench-24~\cite{shen2025finsearch} evaluate open-domain, multi-step, and temporally aware search; and FinDeepForecast~\cite{li2026findeepforecast} considers live financial forecasting.

FinDeepIndicator evaluates financial indicator construction as an end-to-end capability, requiring models to identify formulas, discover reliable sources, collect heterogeneous data, and perform calculations. Unlike existing knowledge, document-QA, and open-world agent benchmarks, it neither provides evidence nor restricts retrieval to a fixed corpus, but evaluates the correctness and completeness of the entire indicator-construction pipeline.

\subsection{Process-based Evaluation}

Process-based evaluation provides finer-grained diagnostics than final-answer evaluation alone. In mathematical reasoning, process supervision assigns correctness labels to individual steps~\cite{lightman2024let}, while ProcessBench~\cite{zheng2025processbench} evaluates whether models can identify the earliest erroneous step. LLM-as-Judge methods, including G-Eval~\cite{liu2023g}, \cite{NEURIPS2023_91f18a12}, and \cite{li2024llms}, provide scalable evaluation for outputs that cannot be assessed reliably through exact matching. Applied at the step level, they can assess local correctness and localize errors. A broader review is provided by \cite{lee2025evaluating}.
Process-oriented evaluation has also been applied to financial reasoning. FinChain~\cite{xie2025finchain} evaluates executable symbolic traces and intermediate consistency, while FinDeepResearch~\cite{zhu2025findeepresearch} evaluates data recognition, metric calculation, and analytical interpretation.

Our evaluation follows the indicator-construction pipeline of formula specification, data collection, and indicator calculation. We use an LLM-as-Judge for semantically equivalent formulas and tolerance-aware numerical comparison for raw data and computed values, allowing errors to be localized to specific stages.

\section{Conclusion}

In this work, we introduced FinDeepIndicator, the first benchmark for evaluating end-to-end financial indicator construction. Covering 234 indicators and 3,350 QA pairs across the U.S. and Chinese markets.
Experiments show that current search-equipped LLMs and DR agents still struggle with this task, particularly in macroeconomic indicators, with data collection emerging as the main bottleneck. We hope FinDeepIndicator will promote the development of more accurate, transparent, and reliable financial agents.

\section{Limitations and Ethical Considerations}

FinDeepIndicator is constructed from publicly available financial and macroeconomic data. We use these data solely for research purposes and avoid collecting sensitive personal information. 

\section{Generative AI Usage}

Generative AI tools were used to assist with code development, language polishing, response parsing, and formula evaluation. 

% ...

\newpage
\bibliographystyle{ACM-Reference-Format}
\bibliography{main}

@book{penman2010financial,
  title={Financial statement analysis and security valuation},
  author={Penman, Stephen H and Penman, Stephen H},
  year={2010},
  publisher={McGraw-Hill/Irwin New York}
}

@book{murphy1999technical,
  title={Technical analysis of the financial markets: A comprehensive guide to trading methods and applications},
  author={Murphy, John J},
  year={1999},
  publisher={Penguin}
}

@book{mishkin2007economics,
  title={The economics of money, banking, and financial markets},
  author={Mishkin, Frederic S},
  year={2007},
  publisher={Pearson education}
}

@inproceedings{shah2022flue,
  title={When FLUE meets FLANG: Benchmarks and large pretrained language model for financial domain},
  author={Shah, Raj and Chawla, Kunal and Eidnani, Dheeraj and Shah, Agam and Du, Wendi and Chava, Sudheer and Raman, Natraj and Smiley, Charese and Chen, Jiaao and Yang, Diyi},
  booktitle={Proceedings of the 2022 Conference on Empirical Methods in Natural Language Processing},
  pages={2322--2335},
  year={2022}
}

@inproceedings{zhu2024benchmarking,
  title={Benchmarking large language models on CFLUE-a Chinese financial language understanding evaluation dataset},
  author={Zhu, Jie and Li, Junhui and Wen, Yalong and Guo, Lifan},
  booktitle={Findings of the Association for Computational Linguistics: ACL 2024},
  pages={5673--5693},
  year={2024}
}

@inproceedings{guo2025fineval,
  title={FinEval: A chinese financial domain knowledge evaluation benchmark for large language models},
  author={Guo, Xin and Xia, Haotian and Liu, Zhaowei and Cao, Hanyang and Yang, Zhi and Liu, Zhiqiang and Wang, Sizhe and Niu, Jinyi and Wang, Chuqi and Wang, Yanhui and others},
  booktitle={Proceedings of the 2025 Conference of the Nations of the Americas Chapter of the Association for Computational Linguistics: Human Language Technologies (Volume 1: Long Papers)},
  pages={6258--6292},
  year={2025}
}

@article{xie2024finben,
  title={FinBen: A holistic financial benchmark for large language models},
  author={Xie, Qianqian and Han, Weiguang and Chen, Zhengyu and Xiang, Ruoyu and Zhang, Xiao and He, Yueru and Xiao, Mengxi and Li, Dong and Dai, Yongfu and Feng, Duanyu and others},
  journal={Advances in neural information processing systems},
  volume={37},
  pages={95716--95743},
  year={2024}
}

@inproceedings{chen2021finqa,
  title={FinQA: A dataset of numerical reasoning over financial data},
  author={Chen, Zhiyu and Chen, Wenhu and Smiley, Charese and Shah, Sameena and Borova, Iana and Langdon, Dylan and Moussa, Reema and Beane, Matt and Huang, Ting-Hao and Routledge, Bryan R and others},
  booktitle={Proceedings of the 2021 Conference on Empirical Methods in Natural Language Processing},
  pages={3697--3711},
  year={2021}
}

@inproceedings{chen2022convfinqa,
  title={ConvFinQA: Exploring the chain of numerical reasoning in conversational finance question answering},
  author={Chen, Zhiyu and Li, Shiyang and Smiley, Charese and Ma, Zhiqiang and Shah, Sameena and Wang, William Yang},
  booktitle={Proceedings of the 2022 conference on empirical methods in natural language processing},
  pages={6279--6292},
  year={2022}
}

@inproceedings{zhu2021tat,
  title={TAT-QA: A question answering benchmark on a hybrid of tabular and textual content in finance},
  author={Zhu, Fengbin and Lei, Wenqiang and Huang, Youcheng and Wang, Chao and Zhang, Shuo and Lv, Jiancheng and Feng, Fuli and Chua, Tat-Seng},
  booktitle={Proceedings of the 59th annual meeting of the Association for Computational Linguistics and the 11th international joint conference on natural language processing (volume 1: long papers)},
  pages={3277--3287},
  year={2021}
}

@article{islam2023financebench,
  title={FinanceBench: A new benchmark for financial question answering},
  author={Islam, Pranab and Kannappan, Anand and Kiela, Douwe and Qian, Rebecca and Scherrer, Nino and Vidgen, Bertie},
  journal={arXiv preprint arXiv:2311.11944},
  year={2023}
}

@inproceedings{chen2024fintextqa,
  title={FinTextQA: A dataset for long-form financial question answering},
  author={Chen, Jian and Zhou, Peilin and Hua, Yining and Xin, Loh and Chen, Kehui and Li, Ziyuan and Zhu, Bing and Liang, Junwei},
  booktitle={Proceedings of the 62nd Annual Meeting of the Association for Computational Linguistics (Volume 1: Long Papers)},
  pages={6025--6047},
  year={2024}
}

@inproceedings{reddy2024docfinqa,
  title={DocFinQA: A long-context financial reasoning dataset},
  author={Reddy, Varshini and Koncel-Kedziorski, Rik and Lai, Viet Dac and Krumdick, Michael and Lovering, Charles and Tanner, Chris},
  booktitle={Proceedings of the 62nd Annual Meeting of the Association for Computational Linguistics (Volume 2: Short Papers)},
  pages={445--458},
  year={2024}
}

@inproceedings{lai2025sec,
  title={SEC-QA: A systematic evaluation corpus for financial QA},
  author={Lai, Viet and Krumdick, Michael and Lovering, Charles and Reddy, Varshini and Schmidt, Craig and Tanner, Chris},
  booktitle={Proceedings of The 10th Workshop on Financial Technology and Natural Language Processing},
  pages={221--236},
  year={2025}
}

@inproceedings{choe2025hierarchical,
  title={Hierarchical retrieval with evidence curation for open-domain financial question answering on standardized documents},
  author={Choe, Jaeyoung and Kim, Jihoon and Jung, Woohwan},
  booktitle={Findings of the Association for Computational Linguistics: ACL 2025},
  pages={16663--16681},
  year={2025}
}

@inproceedings{choi2025finagentbench,
  title={FinAgentBench: A benchmark dataset for agentic retrieval in financial question answering},
  author={Choi, Chanyeol and Kwon, Jihoon and Lopez-Lira, Alejandro and Kim, Chaewoon and Kim, Minjae and Hwang, Juneha and Ha, Jaeseon and Choi, Hojun and Yun, Suyeol and Kim, Yongjin and others},
  booktitle={Proceedings of the 6th ACM International Conference on AI in Finance},
  pages={632--637},
  year={2025}
}

@article{xie2025finchain,
  title={FinChain: A Symbolic Benchmark for Verifiable Chain-of-Thought Financial Reasoning},
  author={Xie, Zhuohan and Orel, Daniil and Thareja, Rushil and Sahnan, Dhruv and Madmoun, Hachem and Zhang, Fan and Banerjee, Debopriyo and Georgiev, Georgi and Peng, Xueqing and Qian, Lingfei and others},
  journal={arXiv preprint arXiv:2506.02515},
  year={2025}
}

@inproceedings{lightman2024let,
  title={Let's verify step by step},
  author={Lightman, Hunter and Kosaraju, Vineet and Burda, Yuri and Edwards, Harrison and Baker, Bowen and Lee, Teddy and Leike, Jan and Schulman, John and Sutskever, Ilya and Cobbe, Karl},
  booktitle={International Conference on Learning Representations},
  volume={2024},
  pages={39578--39601},
  year={2024}
}

@inproceedings{zheng2025processbench,
  title={ProcessBench: Identifying process errors in mathematical reasoning},
  author={Zheng, Chujie and Zhang, Zhenru and Zhang, Beichen and Lin, Runji and Lu, Keming and Yu, Bowen and Liu, Dayiheng and Zhou, Jingren and Lin, Junyang},
  booktitle={Proceedings of the 63rd Annual Meeting of the Association for Computational Linguistics (Volume 1: Long Papers)},
  pages={1009--1024},
  year={2025}
}

@article{lee2025evaluating,
  title={Evaluating Step-by-step Reasoning Traces: A Survey},
  author={Lee, Jinu and Hockenmaier, Julia},
  journal={arXiv preprint arXiv:2502.12289},
  year={2025}
}

@inproceedings{liu2023g,
  title={G-Eval: NLG evaluation using GPT-4 with better human alignment},
  author={Liu, Yang and Iter, Dan and Xu, Yichong and Wang, Shuohang and Xu, Ruochen and Zhu, Chenguang},
  booktitle={Proceedings of the 2023 conference on empirical methods in natural language processing},
  pages={2511--2522},
  year={2023}
}

@inproceedings{NEURIPS2023_91f18a12,
 author = {Zheng, Lianmin and Chiang, Wei-Lin and Sheng, Ying and Zhuang, Siyuan and Wu, Zhanghao and Zhuang, Yonghao and Lin, Zi and Li, Zhuohan and Li, Dacheng and Xing, Eric and Zhang, Hao and Gonzalez, Joseph and Stoica, Ion},
 booktitle = {Advances in Neural Information Processing Systems},
 editor = {A. Oh and T. Naumann and A. Globerson and K. Saenko and M. Hardt and S. Levine},
 pages = {46595--46623},
 publisher = {Curran Associates, Inc.},
 title = {Judging LLM-as-a-Judge with MT-Bench and Chatbot Arena},
 url = {https://proceedings.neurips.cc/paper_files/paper/2023/file/91f18a1287b398d378ef22505bf41832-Paper-Datasets_and_Benchmarks.pdf},
 volume = {36},
 year = {2023}
}

@article{bigeard2025finance,
  title={Finance agent benchmark: Benchmarking LLMs on real-world financial research tasks},
  author={Bigeard, Antoine and Nashold, Langston and Krishnan, Rayan and Wu, Shirley},
  journal={arXiv preprint arXiv:2508.00828},
  year={2025}
}

@inproceedings{hu2026finsearchcomp,
title={FinSearchComp: Towards a Realistic, Expert-Level Evaluation of Financial Search and Reasoning},
author={LIANG HU and Jianpeng Jiao and Jiashuo Liu and Dongyuan Mutu and Yanle Ren and Zhoufutu Wen and Kaiyuan Zhang and Xuanliang Zhang and Xiang Gao and Tianci He and FEI HU and Yali Liao and Zaiyuan Wang and Jingkai Liu and Sun Daibin and Ziqing Zeng and Zhiyuan Zeng and Chenghao Yang and Qianyu Yang and Mingren Yin and Ge Zhang and Xinyi zhang and Xiying ZHAO and Zhu Zhenwei and Hongseok Namkoong and Wenhao Huang},
booktitle={The Fourteenth International Conference on Learning Representations},
year={2026},
url={https://openreview.net/forum?id=8AJbbbe2ni}
}

@inproceedings{shen2025finsearch,
  title={FinSearch: A Temporal-Aware Search Agent Framework for Real-Time Financial Information Retrieval with Large Language Models},
  author={Shen, Yiqing and Zhang, Jingshu and Chen, Feng and Yan, Kaiyuan and Li, Hongguang},
  booktitle={Proceedings of the 6th ACM International Conference on AI in Finance},
  pages={10--17},
  year={2025}
}

@inproceedings{calderon-etal-2025-alternative,
    title = "The Alternative Annotator Test for LLM-as-a-Judge: How to Statistically Justify Replacing Human Annotators with LLMs",
    author = "Calderon, Nitay  and
      Reichart, Roi  and
      Dror, Rotem",
    booktitle = "Proceedings of the 63rd Annual Meeting of the Association for Computational Linguistics (Volume 1: Long Papers)",
    month = jul,
    year = "2025",
    address = "Vienna, Austria",
    publisher = "Association for Computational Linguistics",
    url = "https://aclanthology.org/2025.acl-long.782/",
    doi = "10.18653/v1/2025.acl-long.782",
    pages = "16051--16081",
    ISBN = "979-8-89176-251-0",
}

@article{zhu2025findeepresearch,
  title={FinDeepResearch: Evaluating deep research agents in rigorous financial analysis},
  author={Zhu, Fengbin and Ng, Xiang Yao and Liu, Ziyang and Liu, Chang and Zeng, Xianwei and Wang, Chao and Tan, Tianhui and Yao, Xuan and Shao, Pengyang and Xu, Min and others},
  journal={arXiv preprint arXiv:2510.13936},
  year={2025}
}

@article{li2026findeepforecast,
  title={FinDeepForecast: A Live Multi-Agent System for Benchmarking Deep Research Agents in Financial Forecasting},
  author={Li, Xiangyu and Yao, Xuan and Qi, Guohao and Zhu, Fengbin and Koa, Kelvin JL and Ng, Xiang Yao and Liu, Ziyang and Ni, Xingyu and Liu, Chang and Yang, Yonghui and others},
  journal={arXiv preprint arXiv:2601.05039},
  year={2026}
}

@inproceedings{yao2022react,
  title={ReAct: Synergizing Reasoning and Acting in Language Models},
  author={Yao, Shunyu and Zhao, Jeffrey and Yu, Dian and Du, Nan and Shafran, Izhak and Narasimhan, Karthik R and Cao, Yuan},
  booktitle={The eleventh international conference on learning representations},
  year={2022}
}

@article{li2024llms,
  title={LLMs-as-judges: a comprehensive survey on LLM-based evaluation methods},
  author={Li, Haitao and Dong, Qian and Chen, Junjie and Su, Huixue and Zhou, Yujia and Ai, Qingyao and Ye, Ziyi and Liu, Yiqun},
  journal={arXiv preprint arXiv:2412.05579},
  year={2024}
}

@article{huang2025deep,
  title={Deep research agents: A systematic examination and roadmap},
  author={Huang, Yuxuan and Chen, Yihang and Zhang, Haozheng and Li, Kang and Zhou, Huichi and Fang, Meng and Yang, Linyi and Li, Xiaoguang and Shang, Lifeng and Xu, Songcen and others},
  journal={arXiv preprint arXiv:2506.18096},
  year={2025}
}

@article{zhang2025deep,
  title={Deep research: A survey of autonomous research agents},
  author={Zhang, Wenlin and Li, Xiaopeng and Zhang, Yingyi and Jia, Pengyue and Wang, Yichao and Guo, Huifeng and Liu, Yong and Zhao, Xiangyu},
  journal={arXiv preprint arXiv:2508.12752},
  year={2025}
}

@article{jin2025search,
  title={Search-R1: Training LLMs to reason and leverage search engines with reinforcement learning},
  author={Jin, Bowen and Zeng, Hansi and Yue, Zhenrui and Yoon, Jinsung and Arik, Sercan and Wang, Dong and Zamani, Hamed and Han, Jiawei},
  journal={arXiv preprint arXiv:2503.09516},
  year={2025}
}

@article{song2025r1,
  title={R1-searcher: Incentivizing the search capability in LLMs via reinforcement learning},
  author={Song, Huatong and Jiang, Jinhao and Min, Yingqian and Chen, Jie and Chen, Zhipeng and Zhao, Wayne Xin and Fang, Lei and Wen, Ji-Rong},
  journal={arXiv preprint arXiv:2503.05592},
  year={2025}
}

@online{openai2025introducing,
  title        = {Introducing deep research},
  author       = {OpenAI},
  year         = 2025,
  url          = {https://openai.com/index/introducing-deep-research/},
  urldate      = {2026-7-26}
}

@online{gemini2025deep,
  title        = {Deep Research is now available on Gemini 2.5 Pro Experimental.},
  author       = {Gemini Team},
  year         = 2025,
  url          = {https://blog.google/products/gemini/deep-research-gemini-2-5-pro-experimental/},
  urldate      = {2026-7-26}
}

@online{qwen2025deep,
  title        = {Deep research (Qwen-Deep-Research)},
  author       = {Qwen Team},
  year         = 2025,
  url          = {https://www.alibabacloud.com/help/en/model-studio/qwen-deep-research},
  urldate      = {2026-7-26}
}

@inproceedings{mialon2024gaia,
  title={GAIA: a benchmark for general AI assistants},
  author={Mialon, Gr{\'e}goire and Fourrier, Cl{\'e}mentine and Wolf, Thomas and LeCun, Yann and Scialom, Thomas},
  booktitle={International Conference on Learning Representations},
  volume={2024},
  pages={9025--9049},
  year={2024}
}

@inproceedings{krishna2025fact,
  title={Fact, fetch, and reason: A unified evaluation of retrieval-augmented generation},
  author={Krishna, Satyapriya and Krishna, Kalpesh and Mohananey, Anhad and Schwarcz, Steven and Stambler, Adam and Upadhyay, Shyam and Faruqui, Manaal},
  booktitle={Proceedings of the 2025 Conference of the Nations of the Americas Chapter of the Association for Computational Linguistics: Human Language Technologies (Volume 1: Long Papers)},
  pages={4745--4759},
  year={2025}
}

@article{wei2025browsecomp,
  title={Browsecomp: A simple yet challenging benchmark for browsing agents},
  author={Wei, Jason and Sun, Zhiqing and Papay, Spencer and McKinney, Scott and Han, Jeffrey and Fulford, Isa and Chung, Hyung Won and Passos, Alex Tachard and Fedus, William and Glaese, Amelia},
  journal={arXiv preprint arXiv:2504.12516},
  year={2025}
}

@article{du2025deepresearch,
  title={DeepResearch Bench: A comprehensive benchmark for deep research agents},
  author={Du, Mingxuan and Xu, Benfeng and Zhu, Chiwei and Wang, Xiaorui and Mao, Zhendong},
  journal={arXiv preprint arXiv:2506.11763},
  year={2025}
}

@article{li2025reportbench,
  title={ReportBench: Evaluating deep research agents via academic survey tasks},
  author={Li, Minghao and Zeng, Ying and Cheng, Zhihao and Ma, Cong and Jia, Kai},
  journal={arXiv preprint arXiv:2508.15804},
  year={2025}
}

@article{han2025deer,
  title={DEER: A comprehensive and reliable benchmark for deep-research expert reports},
  author={Han, Janghoon and Kim, Heegyu and Lee, Changho and Lee, Dahm and Park, Min Hyung and Song, Hosung and Jungkyu Choi, Stanley and Lee, Moontae and Lee, Honglak},
  journal={arXiv e-prints},
  pages={arXiv--2512},
  year={2025}
}

@inproceedings{van2019best,
  title={Best practices for the human evaluation of automatically generated text},
  author={Van Der Lee, Chris and Gatt, Albert and Van Miltenburg, Emiel and Wubben, Sander and Krahmer, Emiel},
  booktitle={Proceedings of the 12th international conference on natural language generation},
  pages={355--368},
  year={2019}
}

@inproceedings{elangovan2024considers,
  title={Considers-the-human evaluation framework: Rethinking human evaluation for generative large language models},
  author={Elangovan, Aparna and Liu, Ling and Xu, Lei and Bodapati, Sravan Babu and Roth, Dan},
  booktitle={Proceedings of the 62nd Annual Meeting of the Association for Computational Linguistics (Volume 1: Long Papers)},
  pages={1137--1160},
  year={2024}
}

\appendix
\section{Appendix}

\subsection{Indicator List}
\label{app:indicator}

See Table~\ref{tab:indicator_taxonomy}.

\begin{CJK*}{UTF8}{gbsn}
\begin{table*}[t]
\centering
\setlength{\abovecaptionskip}{0cm}
\setlength{\belowcaptionskip}{0cm}
\caption{Categories, sub-categories, and indicators in FinDeepIndicator.}
\label{tab:indicator_taxonomy}

\scriptsize
\setlength{\tabcolsep}{3pt}
\renewcommand{\arraystretch}{1.05}

\begin{tabular}{
    >{\raggedright\arraybackslash}p{2.0cm}
    >{\raggedright\arraybackslash}p{2.8cm}
    >{\raggedright\arraybackslash}p{12.0cm}
}
\toprule
\textbf{Category} & \textbf{Sub-category} & \textbf{Indicator} \\
\midrule

fundamental
&
profitability
&
Gross Margin, EBIT Margin, EBITDA Margin, Net Profit Margin, Return on Equity, Return on Assets, Return on Invested Capital.
毛利率，EBIT利润率，EBITDA利润率，净利率，净资产收益率，总资产收益率，投入资本回报率。
\\

fundamental
&
operational efficiency
&
Days Inventory Outstanding, Days Sales Outstanding, Days Payable Outstanding, Operating Cycle, Cash Conversion Cycle, Inventory Turnover, Receivables Turnover, Payables Turnover, Total Asset Turnover, Fixed Asset Turnover, Working Capital Turnover.
\newline
存货周转天数，应收账款周转天数，应付账款周转天数，营业周期，现金周转期，存货周转率，应收账款周转率，应付账款周转率，总资产周转率，固定资产周转率，营运资本周转率。
\\

fundamental
&
solvency
&
Current Ratio, Quick Ratio, Cash Ratio, Liability-to-Asset Ratio, Liability-to-Equity Ratio, Net Debt to Equity, Interest Coverage Ratio, Net Debt to EBITDA, Equity Multiplier.
\newline
流动比率，速动比率，资产负债率，产权比率，净债务权益比，利息保障倍数，净债务/EBITDA，权益乘数。
\\

fundamental
&
growth
&
Revenue Growth, EBITDA Growth, Net Income Growth, EPS Growth.
\newline
营业收入增长率，EBITDA增长率，净利润增长率，每股收益增长率。
\\

fundamental
&
cash flow
&
Free Cash Flow, Free Cash Flow to Firm, Free Cash Flow to Equity, Operating Cash Flow Margin, Capex Intensity, Cash Flow to Net Income, Cash Flow to Debt.
\newline
公司自由现金流，股权自由现金流，经营现金流利润率，资本支出占收入比，现金流量充裕率。
\\

fundamental
&
per share
&
Earnings Per Share Basic, Earnings Per Share Diluted, Operating Cash Flow Per Share.
\newline
每股净资产，每股资本公积金，每股未分配利润，每股经营现金流。
\\

fundamental
&
valuation
&
EBIT, EBITDA, NOPAT.
\newline
息税前利润，息税折旧摊销前利润，税后净营业利润。
\\

\midrule

technical
&
moving average
&
Simple Moving Average, Weighted Moving Average, Exponential Moving Average, Double Exponential Moving Average, Triple Exponential Moving Average, Hull Moving Average, Kaufman Adaptive Moving Average.
\newline
简单移动平均，加权移动平均，指数移动平均，双指数移动平均，三指数移动平均，赫尔移动平均，考夫曼自适应移动平均。
\\

technical
&
channel
&
Bollinger Bands, Keltner Channel, Moving Average Envelope, Donchian Channel, SuperTrend.
\newline
布林带，肯特纳通道，包络线，唐安奇通道，超级趋势。
\\

technical
&
trend
&
Aroon Up, Aroon Down, Aroon Oscillator, MACD Line, MACD Signal, MACD Histogram, Parabolic SAR, Percentage Price Oscillator.
\newline
Aroon上升，Aroon下降，Aroon振荡，MACD的DIF，MACD的DEA，MACD，抛物线SAR，百分比价格振荡。
\\

technical
&
momentum
&
Relative Strength Index, Commodity Channel Index, Williams \%R, Stochastic \%K, Stochastic \%D, Stochastic RSI.
\newline
相对强弱指标，商品通道指数，威廉\%R，KDJ的K，KDJ的D，KDJ的J，随机RSI。
\\

technical
&
volatility
&
Bollinger \%B, Average True Range, True Range, Chaikin Volatility.
\newline
布林带\%B，平均真实波幅，真实波幅，蔡金波动率。
\\

technical
&
volume
&
Ease of Movement, Money Flow Index, On-Balance Volume, Force Index, Accumulation/Distribution Line, Chaikin Money Flow, Chaikin Oscillator.
\newline
易动指标，资金流量指数，成交量加权平均价，能量潮，力度指数，累积/派发线，蔡金资金流，蔡金振荡指标。
\\

technical
&
return
&
Annualized Return.
\newline
年化收益率。
\\

technical
&
risk
&
Annualized Volatility, Skewness, Excess Kurtosis, Downside Deviation, Maximum Drawdown, Tracking Error, Ulcer Index, Beta, VaR (Parametric), VaR (Historical Simulation), Expected Shortfall (Parametric), Expected Shortfall (Historical Simulation).
\newline
年化波动率，偏度，超额峰度，下行偏差，最大回撤，跟踪误差，溃疡指数，贝塔系数，VaR（参数法），VaR（历史模拟法），CVaR（参数法），CVaR（历史模拟法）。
\\

technical
&
risk-adjusted return
&
Information Ratio, Sharpe Ratio, Sortino Ratio, Calmar Ratio, Treynor Ratio, Modigliani Ratio, Martin Ratio, Jensen's Alpha.
\newline
信息比率，夏普比率，索提诺比率，卡玛比率，特雷诺比率，莫迪利安尼比率，马丁比率，詹森阿尔法。
\\

\midrule

macro
&
inflation
&
Macro Real Rate, Household Real Rate, Corporate Real Rate.
\newline
GDP平减指数，企业贷款实际利率。
\\

macro
&
productivity
&
Incremental Capital Output Ratio, Total Factor Productivity Growth.
\newline
增量资本产出率，劳动生产率，全要素生产率增长率。
\\

macro
&
external trade
&
Trade Openness.
\newline
贸易开放度。
\\

macro
&
fiscal
&
Federal Debt to GDP, Fiscal Deficit Ratio, Federal Tax Burden.
\newline
中央政府杠杆率，财政赤字率，赤字依存度，宏观税负。
\\

macro
&
labor
&
Real Wage Growth Rate.
\newline
（城镇非私营单位）实际工资增长率，（城镇私营单位）实际工资增长率，就业弹性。
\\

\bottomrule
\end{tabular}

\end{table*}
\end{CJK*}

\subsection{Expert Review Criteria}
\label{app:expert_review_criteria}

Human experts are instructed to evaluate whether the four components extracted by the LLM (the indicator
formula, the raw data table, the indicator calculation table, and
the final answer) faithfully and accurately reflect the information contained in the original model response. Each extracted component is assessed independently according to the following three-level scoring rubric:

\begin{itemize}
    \item \textbf{2 (Correct).} The extracted content is faithful, complete, and compliant with the specified format. It accurately reflects the information explicitly stated in the original response without introducing unsupported or hallucinated content. Numerical precision is properly preserved, and the required unit-conversion rules are followed when applicable. The extraction contains no additional explanations, Markdown syntax, or irrelevant text. When the required information is absent from the original response, the extraction correctly returns \texttt{NA} or leaves the corresponding CSV cell empty, depending on the specification of the field.
    \item \textbf{1 (Partially Correct).} The extracted content captures the main information in the original response but contains minor errors or formatting issues that do not render the extraction unusable. Typical cases include extracting the correct formula with a small amount of extraneous text, omitting a limited number of required CSV cells, or preserving the correct numerical value while retaining undesired formatting such as commas, units, or percentage signs. This category also includes minor errors in numerical precision or unit conversion when the intended value remains clearly identifiable. This score is assigned when the extraction is largely faithful but does not fully satisfy the required format, precision, or completeness constraints.
    \item \textbf{0 (Incorrect).} The extracted content is unfaithful, substantially incomplete, hallucinated, or severely non-compliant with the specified format. Typical cases include extracting a formula, data value, indicator value, or final answer that is not stated in the original response; returning \texttt{NA} when the relevant information is explicitly provided; selecting an intermediate result rather than the final answer; using an incorrect CSV header or column order; fabricating missing values; omitting most of the required content; or producing an output format that makes the extracted result unusable for subsequent evaluation.
\end{itemize}

\subsection{Expert Annotation Rubric for Formula Specification Evaluation}
\label{app:expert_annotation_rubric}

Human experts evaluate the correctness of the formula specified in each model response by comparing the extracted candidate formula with the corresponding reference formula. The resulting score measures whether the candidate formula accurately represents the intended mathematical definition of the target indicator. Experts are instructed to consider both mathematical equivalence and semantic consistency, allowing differences in notation, variable names, and equivalent algebraic forms. Each candidate formula is assigned one of the following scores:

\begin{itemize}
    \item \textbf{2 (Fully Correct).} The candidate formula is mathematically and semantically equivalent to the reference formula. It uses the same required inputs, allowing for synonymous variable names or equivalent notation, and would produce the same numerical result as the reference formula for any valid input.
    \item \textbf{1 (Partially Correct).} The candidate formula captures the general computational structure or identifies the relevant inputs but contains a substantive error that prevents full equivalence with the reference formula. Typical cases include omitting a required term, using an incorrect constant, applying an incorrect aggregation operation such as a sum instead of an average, reversing the numerator and denominator, specifying an incorrect time window, or implementing only part of the required computation.
    \item \textbf{0 (Incorrect).} The candidate formula is incorrect, unrelated to the target indicator, empty, or based on fundamentally different inputs or operations. Such a formula would not compute the intended quantity represented by the reference formula.
\end{itemize}

\end{document}